\documentclass[pdflatex,sn-mathphys]{sn-jnl}

\usepackage{amssymb}
\usepackage{caption}
\usepackage{graphicx}
\usepackage{float} 

\jyear{2021}%

\theoremstyle{thmstyleone}%
\theoremstyle{thmstyletwo}%

\theoremstyle{thmstylethree}%

\begin{document}

\title[S$^{3}$CE]{Self-Supervised Deep Subspace Clustering with Entropy-norm}

\author[1]{\fnm{Guangyi} \sur{Zhao}}\email{gyzhao@hdu.edu.cn}

\author[1]{\fnm{Simin} \sur{Kou}}

\author*[1]{\fnm{Xuesong} \sur{Yin}}\email{yinxs@hdu.edu.cn(X. Yin)}

\author[1]{\fnm{Guodao} \sur{Zhang}}

\author[1]{\fnm{Jieyue} \sur{Yu}}

\author[1]{\fnm{Yigang} \sur{Wang}}

\affil[1]{\orgdiv{Department of Digital Media Technology}, \orgname{ Hangzhou Dianzi University}, \orgaddress{ \city{Hangzhou}, \postcode{310018}, \country{China}}}

\abstract{Auto-Encoder based deep subspace clustering (DSC) is widely used in computer vision, motion segmentation and image processing. However, it suffers from the following three issues in the self-expressive matrix learning process: the first one is less useful information for learning self-expressive weights due to the simple reconstruction loss; the second one is that the construction of the self-expression layer associated with the sample size requires high-computational cost; and the last one is the limited connectivity of the existing regularization terms. In order to address these issues, in this paper we propose a novel model named Self-Supervised deep Subspace Clustering with Entropy-norm (S$^{3}$CE). Specifically, S$^{3}$CE exploits a self-supervised contrastive network to gain a more effetive feature vector. The local structure and dense connectivity of the original data benefit from the self-expressive layer and additional entropy-norm constraint. Moreover, a new module with data enhancement is designed to help S$^{3}$CE focus on the key information of data, and improve the clustering performance of positive and negative instances through spectral clustering. Extensive experimental results demonstrate the superior performance of S$^{3}$CE in comparison to the state-of-the-art approaches.}


\maketitle

\section{Introduction}\label{sec1}

As an important technique of unsupervised learning, clustering is widely used in computer vision, pattern recognition and other fields. It divides the data into different clusters by calculating the similarities among data points and maximizes the inter-cluster variability and intra-cluster similarity. The impressive performance of subspace clustering \cite{bib1,bib3}  in grouping high-dimensional data has made it a hot topic in the field in recent years.

Most subspace clustering focus on spectral clustering-based methods, which are usually divided by two subproblems. One is to construct the affinity matrix based on the self-expression property \cite{bib2}, i.e., each data point can be expressed as a linear combination of other points. The other is to segment data into different clusters via normalized cuts or spectral clustering. In general, constructing a proper affinity matrix is a challenging task. Thus, recent researchers use different types of regularization, such as the nuclear norm \cite{bib4}, $\ell_{0}$-norm \cite{bib7}, $\ell_{2}$-norm \cite{bib5}, traceLasso norm \cite{bib8} and various variants \cite{bib9,bib10,bib11,bib12,bib13,bib14} to get the high-quality affinity matrix. However, it is unsatisfying to simply fit the data into a linear relationship, and it is also difficult to choose a suitable kernel in kernel methods \cite{bib50}.

The enormous success of deep neural networks (DNNs) in many fields gives a new direction to the development of clustering. \cite{bib15,bib16,bib17,bib18,bib19,bib20,bib21,bib23}. Compared with traditional subspace clustering, deep subspace clustering methods utilize deep neural networks to learn a nonlinear mapping from the original data samples to a latent feature space. The introduction of DNNS compensates for the lack of attention to a local structure by the self-expression property. Various deep models such as Autoencoders (AE) \cite{bib25}, Generative Adversarial Network (GAN) \cite{bib26} and Variational Autoencoders (VAE) \cite{bib27} are used in deep subspace clustering to obtain a good representation of the data. An impressive model is deep subspace clustering (DSC) \cite{bib16}, which introduces a self-expression layer between the encoder and decoder in the autoencoder. Although the emergence of the AE brings a qualitative leap in the effectiveness of subspace clustering, the quality of the representations obtained by it is beyond satisfactory. In other words, the reconstruction loss of AE focuses too much on pixel-level differences and ignores important local information between neighboring points. Deep subspace clustering expects to obtain a representation through the model that better reflects the important features of the data, but reconstruction loss imposes a resistance to this goal.

Self-supervised learning is a kind of the unsupervised learning, which learns a general feature representation for downstream tasks by supervising itself. Traditional autoencoders mainly obtain a representation of the original data sample in the implicit space by reconstructing the paradigm of the input data \cite{bib25}. This representation, although being effective from the point of view of data reduction, is more redundant for discriminative tasks, e.g., clustering and classification. The reason is that it gives the same level of attention to various implicit properties in the representation rather than focusing on the most representative aspects of the data, which is inefficient for discriminating data. In contrast, contrastive learning \cite{bib28}, based on the idea of maximizing mutual information, is dedicated to using fewer symbols to express more critical information. That is, it concentrates more on patterns that occur frequently or repeatedly in the data samples. This reduces the redundancy and improves the efficiency and friendliness of representation learning for discriminative tasks. The core idea of contrastive learning is to acquire a representation learning model by constructing similar and dissimilar instances. Then a representation is obtained that correctly selects positive examples from a dataset of mixed positive and negative example samples. The representation is able to uncover the essence of the data and then discriminate it, which facilitates the efficient completion of downstream tasks.

\begin{figure}[!t]
\centering
\includegraphics[scale=0.5]{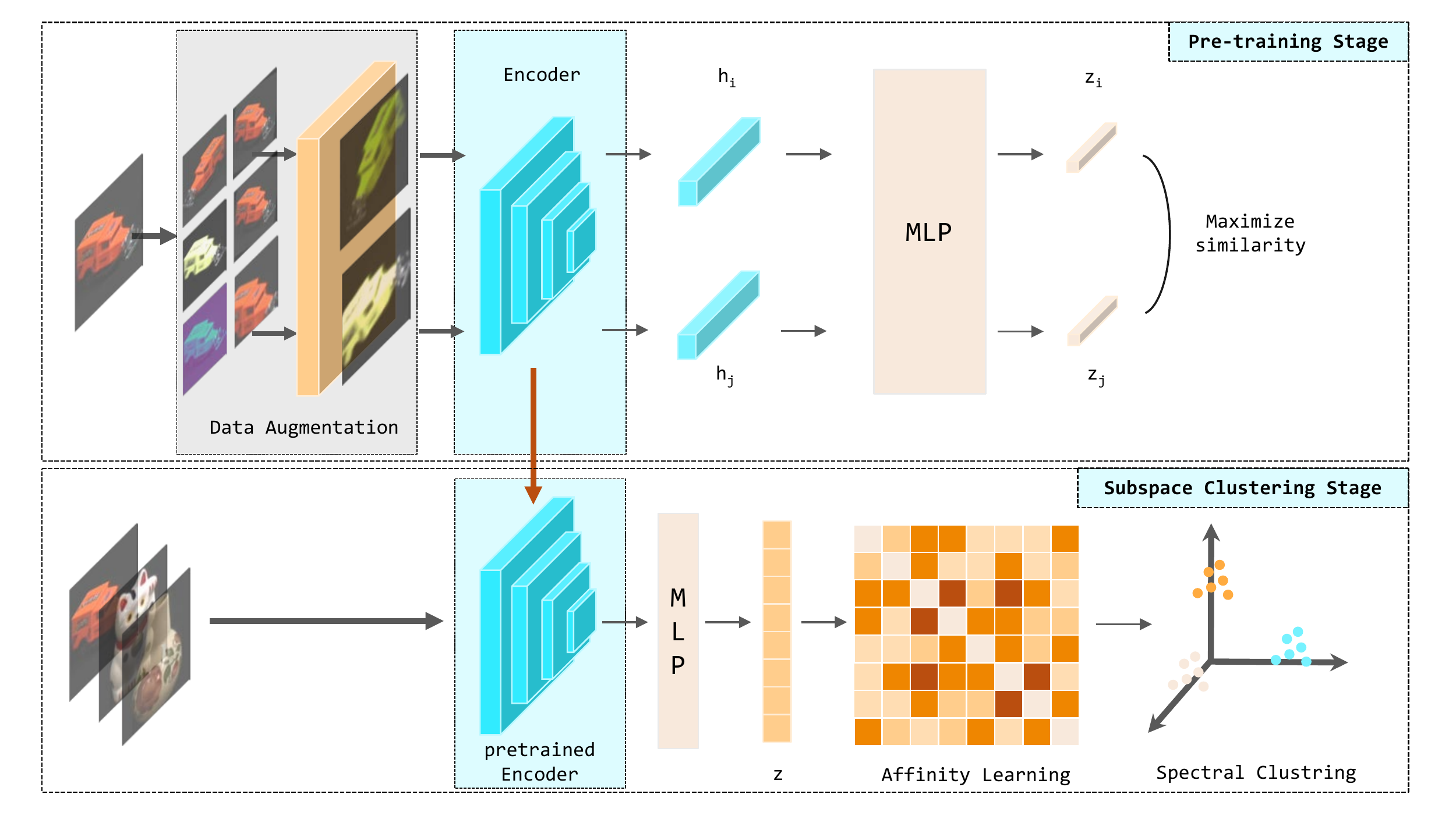}
\caption{The architecture of Self-Supervised Deep Subspace Clustering with Entropy-norm (S$^{3}$CE ). S$^{3}$CE  consists of two stages: pre-training stage and subspace clustering stage. The data augmented is used to train an encoder by self-supervised contrastive learning in the pre-training stage. In particular, S$^{3}$CE  feeds the data into a pre-trained encoder and fine-tunes it. The self-expression property and the entropy-norm are used to obtain the clustering results}
\label{fig:1}
\end{figure}

Motivated by recent progress in deep subspace clustering and self-supervised learning, in this paper we propose a self-supervised deep subspace clustering with entropy-norm(S$^{3}$CE). S$^{3}$CE can obtain a good encoder by contrastive discriminative methods during the pre-training stage, and then the encoder will be used to get representation with frozen parameters. Finally, our method will use the representation after the projection layer mapping to obtain the similarity matrix with entropy-norm. The result on seven benchmark datasets show the effectiveness of our method.

As shown in Fig. \ref{fig:1}, the structure of our method consists of a pre-training phase and a deep subspace clustering phase. The highlights of our approach are the following four-fold:

(1) Rather than focusing on self-reconstruction which tends to be detrimental to the relationships between data points, we utilize one encoder to reduce the time cost in the learning process. In addition, the local structure information between data points is effectively preserved after removing the encoder.

(2) We introduce the entropy-norm for better constraining the affinity to unify and tension the distribution of data samples in each subspace, which is more useful to improve the performance with the next step of the spectral clustering.

(3) We exploit data augmentation to allow the encoder to learn the key information in the data and fine-tune the frozen model with two linear layers. Moreover, we present a general framework by combining contrastive learning into DSC, which is more suitable for solving real-world problems.

(4) We analyze the consistency of contrast learning and the entropy-norm in constructing and optimizing
the coefficient matrix from an entropy perspective. It is important to form a clear, uniform and well-spaced block-diagonal matrix.

\section{Related Works}\label{sec2}

\subsection{Traditional Subspace Clustering}\label{subsec2}

Subspace clustering (SC) refers to the method of dividing the data into different classes, and each class corresponds to a different subspace under ideal conditions. Most traditional SC algorithms are based on linear representation, namely self-expression theory. Specifically, the whole sample data matrix is used as a dictionary to obtain the linear representation of each data point (e.g. Eq. \ref{con:eq2}), and then the similarity matrix $W$ is constructed by using the obtained coefficient matrix C. Mathematically, this idea can be expressed as:

\begin{equation}
\min _{C \in \mathbb{R}^{n \times n}} \frac{1}{2}\|X-X C\|_{F}^{2}+\|C\|_{p} \quad \text { s.t. } \operatorname{diag}(C)=0\label{con:eq1}
\end{equation}
where $\|\cdot\|_{F}^{2}$ represents Frobenius norm, $X$ is the data matrix consists of a set of data points $\left\{x_{i}\right\}_{i=1, \ldots, N}$ which let the data points belong to the union of $K$ linear subspace $\left\{S_{i}\right\}_{i=1}^{K}$. Here $\|\cdot\|_{p}$ represents any one norm, and uses the diagonal constraint on $C$ avoids trivial solutions for sparsity inducing norms, such as the $\ell_{1}$-norm in SSC \cite{bib2}, the $\ell_{2}$-norm in LSR \cite{bib11}, the $\ell_{2,1}$-norm in LRR \cite{bib4}, the nuclear norm in LRSC \cite{bib29}, the Forbenius norm in EDSC \cite{bib30}, etc. It is noteworthy that the properties of data similarity matrix $W$ are sparse, symmetrical, nonnegative and nonlinearly representational. In fact, the SSC algorithms hardly guarantee the nonnegative and symmetry attribute of matrix $W$ because SSC explicitly converts the coefficient matrix $C$ by Eq. \ref{con:eq3} into a symmetrical and nonnegative matrix $W$. This conversation may bring some misleading information. Although some works add a few mandatory constraints such as the symmetric positive semi-definite (PSD) \cite{bib31}, the computation of sparse optimization program is very complicated, and it may even be a NP-Hard problem.
\begin{equation}
x_{j}=\sum_{i \neq j} c_{i j} x_{i}+e_{j}\label{con:eq2}
\end{equation}
where $e_{j}$ denotes as a disturbance term for noise or corruption, $x_{j}$ is an arbitrary data point, $c_{ij}$ represents to an entry of column $j^{th}$ of row $i^{th}$ in self-expression coefficient matrix $C$. Moreover, $c_{ij} \neq 0$ and $c_{i i}=0$ are guaranteed to maintain the block-diagonal structure under the assumption of subspace independence \cite{bib30}.
\begin{equation}
W=\frac{(\lvert C \rvert+\lvert C^{T}\rvert)}{2}\label{con:eq3}
\end{equation}

Specifically, SC relies on the self-expressive to obtain coefficient matrix $C$, then generally constructs the data similarity graph $W$ by Eq. \ref{con:eq3}, and finally leverages the spectral clustering to obtain the clustering results.

\subsection{Deep Subspace Clustering}\label{subsec2}

Compared with traditional subspace clustering, deep subspace clustering (DSC) aims to seek the deep representation of data samples. Deep representation of the original data is used to optimize the affinity matrix by a typical sparse or low-rank regularization strategy. This substantially improves the performance of subspace clustering for high-dimensional complex data samples at the nonlinear structure level. The well-known DSC-Net \cite{bib16} with the addition of an autoencoder describes the generic DSC model as:
\begin{equation}
\mathcal{L}(\Theta, C)=\frac{1}{2}\|X-\hat{X}\|_{F}^{2}+\lambda_{1}\|C\|_{q}+\lambda_{2}\|Z-Z C\|_{F}^{2}, \quad \text { s.t. } \operatorname{diag}(C)=0\label{con:eq4}
\end{equation}
where $\hat{X}, Z$ represents the original data $X$ reconstructed by the decoder and the output of the encoder. $\lambda_{1}>0$ and $\lambda_{2}>0$ are the tradeoff parameters. DSC-Nets, as a general framework for DSC, introduces a self-representation layer constructed by a single fully connected network layer without bias and activation to implement the self-representation property in traditional subspace clustering. It is ingenious that embeds the layer between the encoder and the decoder to guide the encoder to learn a valid discriminative representation.

Following the DSC framework, a number of works have been proposed. DASC \cite{bib18} gets the real data and the faked data after a linear combination by the generator and feeds them into the discriminator. After adversarial learning, a better similarity matrix $C$ and feature expression $Z$ would be obtained. Using a dual self-supervision mechanism, S2ConvSCN \cite{bib20} makes the convolution, self-representation module and spectral clustering into an organic whole. Also, for self-supervised learning, PSSC \cite{bib23} introduces pseudo-graphs and pseudo-labels to supervise similarity learning during the training process. In addition, DSC is also used in multi-view \cite{bib33} and multi-modal data \cite{bib32}. However, almost all existing methods use the self-reconstruction loss which is less friendly to relationships between data points.

\subsection{Self-Supervised Learning}\label{subsec2}
According to Liu et al. \cite{bib48}, self-supervised learning is divided into three main categories: Generative, Contrastive, and Adversarial (Generative-Contrastive). Generative methods mainly map the original data to an explicit vector and then revert to the original data through losses such as Mean Square Error (MSE). The representative of this type of approach is Auto-Encoding (AE) model. The adversarial approach improves the realism of the generated images using the generator against the discriminator. The GAN model is a concrete manifestation of this method.

Contrastive learning focuses on salient feature areas of an image by pulling in positive examples and pushing out negative examples. Unlike mapping features to classes singularly, contrastive learning increases inter-class dispersion and intra-class compactness. With the information entropy loss constraint, the model can learn valid class-related information, which guides more robust feature learning.

Proposed by Oord et al. \cite{bib28}, the objective function of the earliest contrastive self-supervised models Contrastive Predictive Coding (CPC) can be described by Eq. \ref{con:eq5} called InfoNCE. To learn better representations, self-supervised research work based on the comparison of positive and negative examples has developed rapidly in recent years. Tian et al. \cite{bib35} proposed Contrastive Multiview Coding (CMC), which increases the number of positive examples with multimodality of data and uses a cached form of Memory Bank to quickly acquire more negative example representations. However, the representation is not a real-time negative example representation acquired by the current model and the information lacks consistency. Thereafter, He et al. \cite{bib36,bib37,bib38} proposed the MoCo model using a dictionary-based momentum update method, which solved the non-consistency problem of CMC to some extent. Focusing more on visual content, Chen et al. \cite{bib39} proposed a comparative representation learning framework SimCLR. SimCLR treats a pair of augmented data of a certain data sample as a positive example and augmented data of other samples as a negative example. It is inspiring to design a general framework and achieve acceptable results at the same time.
\begin{equation}
\mathcal{L}_{N}^{\text {InfoNCE }}=-\mathbb{E}_{X}\left[\log \frac{\exp \left(f(x)^{T} f\left(x^{+}\right)\right)}{\exp \left(f(x)^{T} f\left(x^{+}\right)\right)+\sum_{j=1}^{N} \exp \left(f(x)^{T} f\left(x^{-}\right)\right)}\right]\label{con:eq5}
\end{equation}
where $f(\cdot)$ is the model to be learned. $x^{+}$  and $x^{-}$ denote positive and negative examples respectively, i.e. , similar samples and dissimilar samples.

\section{Proposed Method}\label{sec3}
In this section, we will present our S$^{3}$CE in detail. Our model contains two main parts: a contrastive pre-training module and a subspace clustering module with entropy-norm. In addition, we will discuss the entropy-norm regularized DSC. 

\subsection{Pre-training stage with Contrastive Learning}\label{subsec3}

In general, we assume that the more important information of the original data is retained, the better the feature extraction is. One conventional practice is that reducing the encoded vector back to the original image through the decoder. The final loss is the MSE of the original image and the reconstructed image. Later, the distribution of the encoded vector is limited to be as close to a Gaussian distribution as possible, which leads to the Variational Autoencoders. However, one can find that the reconstruction of the original image by low-dimensional encoding is usually blurred, which can be explained by the fact that the loss function MSE requires too demanding pixel-by-pixel reconstruction. Alternatively, it can be explained by the absence of a suitable loss for image reconstruction. The ideal approach would be to train a discriminator using an adversarial network, but this would further increase the difficulty of the task.

But the truth is that when dealing with tasks like image classification, image clustering, models don't need to draw these images all over again but just distinguish the differences in these images. That is, reasonable and sufficient features for the dataset and task do not necessarily accomplish image reconstruction. The basic principle of a good feature should be the ability to identify that sample from the entire data set. In other words, extracting the (most) unique information about the sample. How to measure that the extracted information is unique to the sample? We use Eq. \ref{con:eq5} proposed by Oord et al. \cite{bib28} instead of MSE calculated by the encoder and decoder to measure.

In the pre-training stage, S$^{3}$CE first obtains $2N$ enhanced images after a combination of several data enhancement operations such as random cropping, random rotation, random color jitter, Gaussian Blur, random grayscale and flipping on a batch of $N$ input images. These data enhancements methods are randomly combined, which can introduce more uncertainty to the data to be fed into the model. Next, the $2N$ enhanced images are put into the encoder to obtain $N$ pairs of high-dimensional vector $h_{a}$ and $h_{b}$. Then, $h_{i}$ and $h_{j}$ will pass through a projection layer consisting of MLPs into $N$ pairs of $z_{i}$ and $z_{j}$. Finally, we compute the cosine similarity of each pair $z_{i}$ and $z_{j}$ as $s_{i,j}$.
\begin{equation}
s_{i, j}=\frac{z_{i} z_{j}}{\left\|z_{i}\right\|\left\|z_{j}\right\|}\label{con:eq6}
\end{equation}

A positive pair of the infoNCE loss $l(i,j)$ is defined as:
\begin{equation}
l(i, j)=-\log \frac{\exp \left(\frac{s_{i, j}}{T}\right)}{\sum_{k=1}^{2 N} \exp \left(\frac{s_{i, k}}{T}\right)} \quad(k \neq i)\label{con:eq7}
\end{equation}
where $T$ denotes a temperature parameter. The final loss of the pre-training stage is calculated for all positive pairs, which can be defined as:
\begin{equation}
\mathcal{L}_{1}=\frac{1}{2 N} \sum_{k=1}^{N}[l(2 k-1,2 k)+l(2 k, 2 k-1)]\label{con:eq8}
\end{equation}

S$^{3}$CE chooses Resnet-50 \cite{bib42} as encoder to handle the images processed by data enhancement. Logically, there is no requirement for the size of the dataset at this stage. We map the feature vector of the encoder output to $B\times N$ dimensions, where $B$ is the batch size and $N$ is related to the chosen encoder. The similarity is calculated using a two-layer MLPs to project the features into the lower dimension. Pre-trained encoder will freeze the weight for the next stage as a feature extractor.

\subsection{Entropy-norm regularized DSC}\label{subsec3}

In reinforcement learning, the probability of the output sometimes is expected not to be focused on one action when training a neural network . At least some non-zero probability should be given to other actions that can be explored. The entropy component can be added to the final loss to make the probability distribution more chaotic, that is, to make the probability distribution more uniform and not simply over-converge to one action. For example, the purpose of SAC \cite{bib49} is to encourage exploration and achieve a multi-modal effect.

High-quality subspace clustering requires high similarity within subspaces and low similarity between subspaces, i.e., tight aggregation within subspaces and mutual independence of each subspace. Among them, there may be multiple connected components in each subspace under the sparse constraint. That is, the weak connection between certain data within that subspace increases the possibility of these samples being incorrectly partitioned into adjacent subspaces. While there is only one connected component in each subspace under the low-rank constraint, some data points are only associated with a small number of samples in that subspace, which leading to a larger possibility that these data samples are incorrectly assigned to other subspaces.

\begin{figure}[!t]
\centering
\includegraphics[scale=0.38]{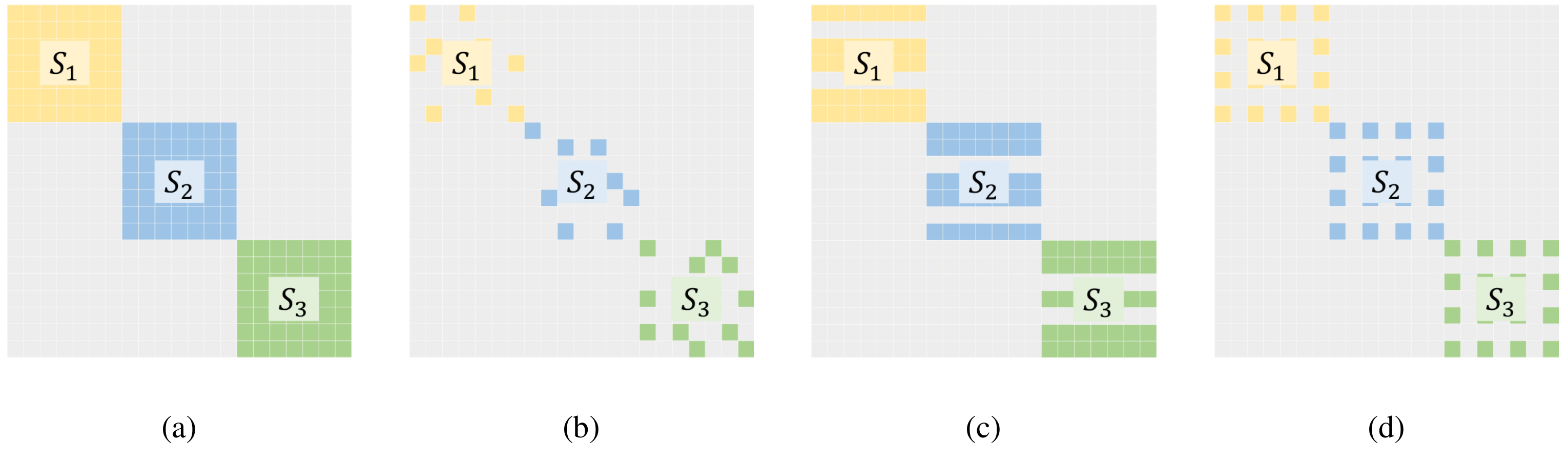}
\caption{Schematic plot of various affinity matrices under different norm constraints. (a)Ideal (b)Sparse (c)Low-rank (d)Ours}
\label{fig:2}
\end{figure}

Entropy-norm was proposed by Bai et al. \cite{bib34} as a regularization term to replace the $\ell_{1}$-norm for SSC, namely SSC+E, and the objective function is defined as:
\begin{equation}
\mathrm{Q}=\sum_{i=1}^{n} \sum_{j=1}^{n}\left\|x_{i}-x_{j}\right\| w_{i j}+\gamma \sum_{i=1}^{n} \sum_{j=1}^{n} w_{i j} \ln w_{i j} \quad \text { s.t., } \sum_{i=1}^{n} w_{i j}=1, \operatorname{diag}(W)=0\label{con:eq9}
\end{equation}
where $w_{ij}$ represents the $i^{th}$ row and $j^{th}$ column of the affinity matrix $W$. Under the regularization of entropy-norm, the similarity between data in the subspace is uniformly and densely distributed, which is more suitable for the requirement of tight aggregation in the subspace. This distribution is easier to achieve concatenation between any two data samples. The schematic plots of the affinity matrices with different norm constraint is shown in Fig. \ref{fig:2} \cite{bib51}. One can observe that the affinity matrix used for spectral clustering has better connectivity under the entropy-norm constraint. Its uniformly distributed block-diagonal structure is closer to the affinity matrix in the ideal state.

\subsection{Subspace clustering stage with Entropy-norm}\label{subsec3}

Inspired by SSC+E and the power of DSC-Net, we impose the entropy-norm constraint on the coefficient matrix $C$ learned from the encoder. Meanwhile, considering that reconstruction loss not only ignores the rich relational information between adjacent points, but also reduces the quality of potential representations. Our model only use the self-expression loss as well as the entropy-norm constraint to train the coefficient matrix. Considering that $c_{ij}$ is a value between 0 and 1, the value after the entropy-norm constraint is negative. Therefore, we use a function to constrain the value of this item to between 0 and 1. The object function of the subspace clustering module can be defined as:
\begin{equation}
\mathcal{L}_{2}=\lambda_{1}\|Z-Z C\|_{F}^{2}+\lambda_{2} \exp \left(\sum_{i=1}^{n} \sum_{j=1}^{n} c_{i j} \ln c_{i j}\right), \quad \text { s.t. } \sum_{j=1}^{n} c_{i j}=1\label{con:eq10}
\end{equation}
where $Z$ represents the output of the encoder obtained from the pre-training stage. $\lambda_1>0$ and $\lambda_2>0$ are the tradeoff parameters. $c_{ij}$ denotes the $i^{th}$ row and $j^{th}$ column of the coefficient matrix $C$. Unlike most existing deep subspace clustering algorithms that use a typical method to enhance the block-diagonal structure \cite{bib30}. After the coefficient matrix $C$ obtained, S$^{3}$CE uses Eq. \ref{con:eq4} to calculate the affinity matrix W directly. Compared to previous methods, the proposed method can eliminate the need to explore multiple trade-off parameter. It makes our method more general and can be applied to more practical problems.

Therefore, the objective functions \ref{con:eq8} and \ref{con:eq10} consist the entire objective functions of our S$^{3}$CE.
Mutual information is applied to contrastive learning from the perspective of entropy. Mutual information is a measure of the correlation between two sets of events. In other words, it is the amount of information obtained by observing another random variable after obtaining information about one random variable.  The potential variables shared by the input are extracted in CPC \cite{bib28} by maximizing the mutual information between the coded representations.

The use of entropy-norm in subspace clustering defines the maximum information entropy as the regular term driving the learning affinity matrix W. The problem of poor symmetry of the sparse self-representation matrix can be solved.

Contrastive learning based on mutual information in the pre-training stage reduces the similarity between each potential vector except itself. It makes the representations encoded by the encoder enhanced independence between each potential vector when constructing the self-expression matrix. Moreover, the weights between different categories in the coefficient matrix are reduced. At the same time, thanks to the entropy-norm, each sample occupies neither too large nor too small a weight in the coefficient matrix when the same categories are represented to each other in the coefficient matrix. The distribution of the data samples within each subspace are more uniform and dense. This benefits to the next step of spectral clustering to perform graph partitioning algorithm in accuracy, which improves the subspace clustering performance.

\section{Experiments}\label{sec4}

To evaluate the effectiveness of our S$^{3}$CE, we conduct experiments on seven benchmark datasets and compare with eight subspace clustering methods in this section.

\subsection{Datasets}\label{subsec4}

\begin{figure}[!t]
\centering
\includegraphics[scale=0.42]{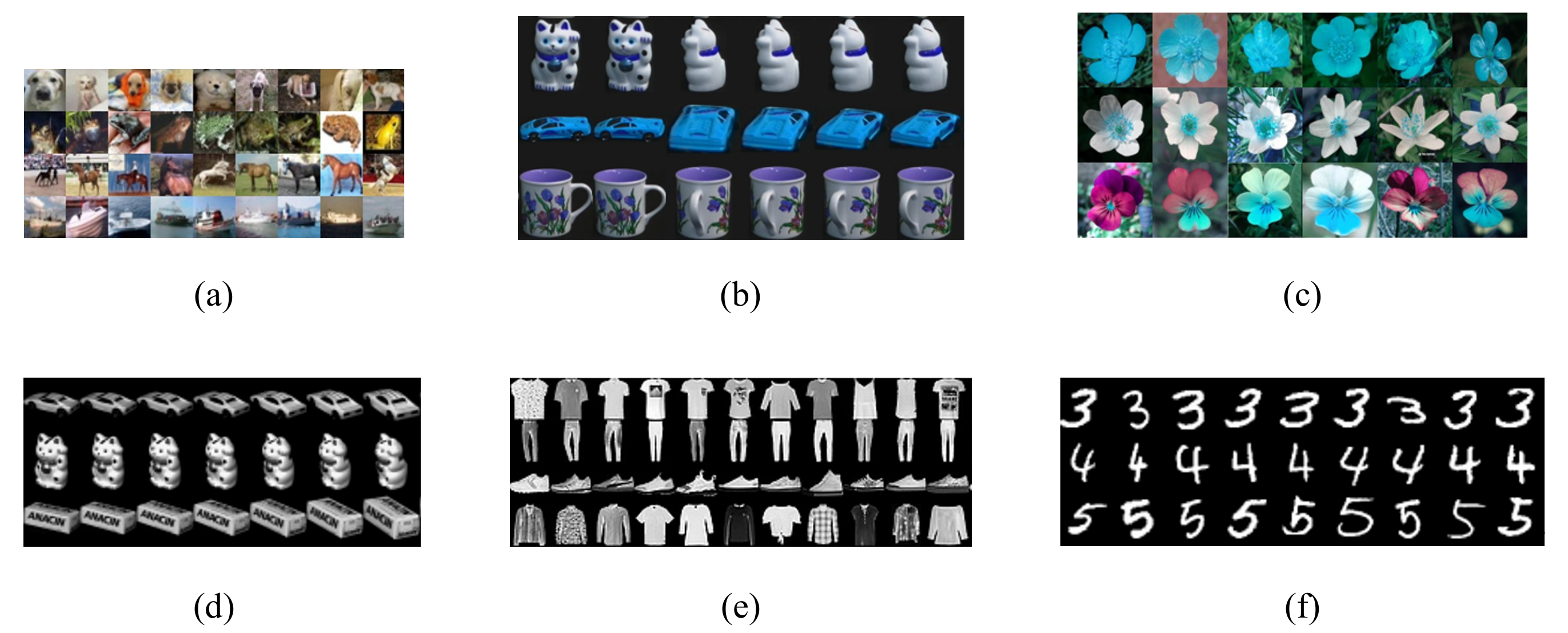}
\caption{Sample images from seven datasets used in the experiments. (a) CIFAR10 (b) COIL40 and COIL100 (c) OxFlowers17 (d) COIL20 (e) Fashion-MNIST (f) MNIST-1000}
\label{fig:3}
\end{figure}

Datasets include three gray-scale datasets (COIL20 \cite{bib41}, Fashion-MNIST \cite{bib44} and MNIST-1000 \cite{bib43}) and five RGB datasets (COIL40 \cite{bib41}, COIL100 \cite{bib41}, CIFAR-10 \cite{bib45}, OxFlowers17 \cite{bib46}).
\begin{itemize}
    \item \textbf{COIL20:} This dataset contains 1,440 gray-scale images of 20 objects with 72 different views. The original image is 128 × 128 pixels. In order to construct positive examples, we resize it to 64 × 64 pixels.
    \item \textbf{Fashion-MNIST-test:} We use the test part of the dataset Fashion-MNIST. Fashion-MNIST-test covers 10,000 different front images of products from 10 categories (e.g., coat, trouser, shirt, dress, bag, etc.), where each gray-scale image is of size 28 × 28.
    \item \textbf{MNIST-1000:} MNIST consists of 70,000 handwritten digits (0-9) in the gray-scale level. We randomly selected 100 images from each of these 10 categories to form Mnist-1000. And each image has 28 × 28 pixels.
    \item \textbf{COIL40, COIL100:} Like coil20, coil40 and coil100 are datasets consisting of different objects imaged at different angles in a 360° rotation, containing 128 × 128 color images of 100 objects (each with 72 poses). Different from COIL100, the first 40 classes of COIl100 are taken as COIl40. We also resize them to 64×64 pixels.
    \item \textbf{CIFAR-10:} CIFAR-10 is a dataset for identifying pervasive objects. A total of 10 categories of RGB color images are included (e.g. automobile, bird, cat, deer, etc.). Each image in this dataset has 32×32 pixels. CIFAR-10 is divided into CIFAR-10-train and CIFAR-10-test, consisting of 50,000 and 10,000 images, respectively. We use the CIFAR-10-test for training.
    \item \textbf{OxFlowers17:} This dataset contains 17 species of flowers selected by the University of Oxford that are common in the UK. The whole dataset has 1360 images and each specie has 80 images. We resize each image to the pixel size of 64 × 64.
\end{itemize}

The detailed information for the datasets are summed up in Table \ref{tab1} and the examples from seven datasets are shown in Fig. \ref{fig:3}.

\begin{table}[h]
\begin{center}
\begin{minipage}{174pt}
\caption{Statistics Of The Datasets}\label{tab1}%
\begin{tabular}{@{}llll@{}}
\toprule
Dataset & Samples  & Classes & Dimensions\\
\midrule
COIL20	& 1440	& 20 & 64×64\\
COIL40	& 2880	& 40 & 64×64\\
COIL100	& 7200	& 100 & 64×64\\
OxFlowers17 & 1360 & 17 & 64×64\\
MNIST-1000 & 1000 & 10 & 28×28\\
CIFAR-10 & 10000 & 10 & 32×32\\
Fashion-MNIST & 10000 & 10 & 28×28\\
\botrule
\end{tabular}
\end{minipage}
\end{center}
\end{table}

\subsection{Comparison Methods}\label{subsec4}
For fair evaluation and comparison, we compare S$^{3}$CE with both traditional and deep subspace clustering methods, including Sparse Subspace Clustering (SSC) \cite{bib2}, elastic net subspace clustering (ENSC) \cite{bib47}, SSC by orthogonal matching pursuit (SSC-OMP) \cite{bib22}, Low-rank representation (LRR) \cite{bib4}, efficient dense subspace clustering (EDSC) \cite{bib30}, DSC-Nets \cite{bib16}, Deep Adversarial Subspace Clustering (DASC) \cite{bib18}, Self-Supervised Convolutional Subspace Clustering Network (S2ConvSCN) \cite{bib20}.

\subsection{Evaluation Metrics}\label{subsec4}
To evaluate the clustering performance of our approach, we use the standard unsupervised evaluation metrics for comparing the results with the above methods, which are unsupervised clustering Accuracy rate (ACC) and Normalized Mutual Information (NMI). The higher value of ACC and NMI indicates the better performance they achieved. ACC can be defined as:
\begin{equation}
ACC=\max _{m} \frac{\sum_{i-1}^{n} \mathbb{I}\left\{l_{i}=m\left(c_{i}\right)\right\}}{n}\label{con:eq11}
\end{equation}
where $l_{i}$ is the ground-truth label, $c_{i}$ denotes the subspace clustering assignment generated by the algorithm, i.e., the number of clusters K, and m is the range of all possible one-to-one mappings between subspace clustering and labels. Furthermore, these mappings are usually computed by Hungarian algorithm in most of existing implementation. NMI represents the correlation between predict labels and ground-truth labels, where NMI ranges from 0 to 1. The discrete calculation of NMI is according to:
\begin{equation}
NMI=\frac{\sum_{i=1}^{K} \sum_{j=1}^{K} n_{i j} \log \left(\frac{n \cdot n_{i j}}{n_{j} \cdot \hat{n}_{j}}\right)}{\sqrt{\left(\sum_{i=1}^{K} n_{i} \log \frac{n_{i}}{n}\right)\left(\sum_{j=1}^{K} \hat{n}_{j} \log \frac{\hat{n}_{j}}{n}\right)}}\label{con:eq12}
\end{equation}
where $n_{i}$ denotes the sample number of the $c_{i} (1 \leq i \leq K)$ cluster provided by the clustering algorithm, $\hat{n}_{j}$ is the number of data belonging to the $j^{th}$ ground-truth class $K$, and $ n_{ij}$ is the number of data that are in the intersection between the cluster $c_{i}$ and the $j^{th}$ class.

\subsection{Implementation Details}\label{subsec4}
We use the Adam optimizer with momentum by the gradient descent algorithm to minimize the loss function, where the learning rate in pre-training stage is set as $3.0 \times 10^{-4}$ and the learning rate in subspace clustering stage is set as $1.0 \times 10^{-5}$. About hyperparameters, we set $\lambda_{1} = 1.0$ and $\lambda_{2} = 75.0$ for COIL20 and COIL40, $\lambda_{1} = 1.0$ and $\lambda_{2} = 15.0$ for COIL100, $\lambda_{1} = 0.1$ and $\lambda_{2} = 10.0$ for OxFlowers17, $\lambda_{1} = 1.0$ and $\lambda_{2} = 6.0$ for OxFlowers17, It is noteworthy that in the pre-training stage with contrastive learning, the larger batch size the higher quality of encoder feature extraction proved by Chen et al. \cite{bib39}. To be fair, in the pre-training stage, the batch size of all datasets are set to 256. Furthermore, the deep auto-encoder based on backbone network Resnet-50 \cite{bib42} is designed in this paper, but our method is non-limited to this convolutional network. It can also be a fully-connected or another convolutional auto-encoder. We also implemented the baseline methods, in which the details strictly follow the source codes released by the authors or the suggestions from the corresponding paper. Finally, it should be stressed here that all the experiments are implemented in Python using PyTorch1.7.1 on an NVIDIA GeForce RTX 3060 GPU.

\subsection{Results}\label{subsec4}
Table \ref{tab2} shows the clustering results on the seven publicly available datasets. The top-ranked method in each measure is highlighted in bold. In most cases, S$^{3}$CE performs significantly better than other methods. Particularly, we have the following observations.

\begin{figure}[!t]
\centering
\includegraphics[scale=0.5]{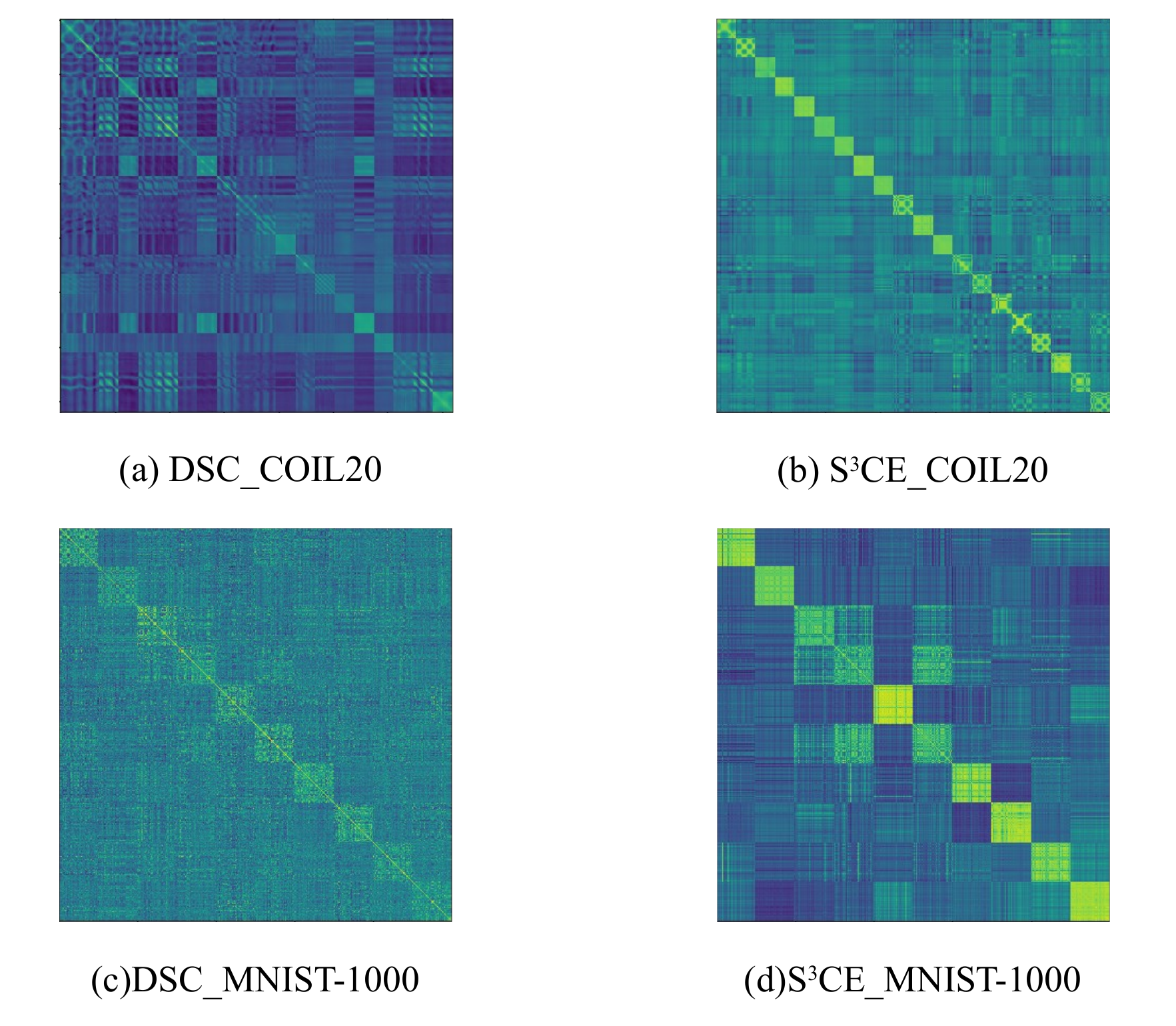}
\caption{Visualization of the affinity matrices for all subjects on COIL20 and MNIST-1000 dataset calculated from Eq. \ref{con:eq4}}
\label{fig:4}
\end{figure}

\begin{itemize}
    \item S$^{3}$CE performs outstandingly better than traditional linear subspace clustering methods. This is mainly attributed to the excellent fitting ability of the DNN model.
    \item S$^{3}$CE outperforms the other deep clustering methods including DSC. Specially, S$^{3}$CE extracts more representative features through comparative learning and focuses more on local information on object datasets. We have also tried to test on the face dataset but found the results to be poor. The reason for this result is speculated to be that the feature differences between faces are not obvious. It is difficult for contrastive learning to learn effective information from them.
    \item Compared with  DASC \cite{bib18} and S2ConvSCN \cite{bib20} ,our S$^{3}$CE produce superior performance . Concerning S2ConvSCN, the performance of DASC is more stable.
\end{itemize}

\begin{sidewaystable}[htbp]
\small
  \centering
  \caption{Clustering Results On Benchmark Datasets}\label{tab2}
   \setlength{\tabcolsep}{0.8mm}{
    \begin{tabular}{lllllllllllllll}
    \toprule
    Datasets & \multicolumn{2}{l}{COIL20} & \multicolumn{2}{l}{COIL40} & \multicolumn{2}{l}{COIL100} & \multicolumn{2}{l}{OxFlowers17} & \multicolumn{2}{l}{MNIST-1000} & \multicolumn{2}{l}{CIFAR-10} & \multicolumn{2}{l}{Fashion-MNIST} \\
    \midrule
    Methods & ACC   & NMI   & ACC   & NMI   & ACC   & NMI   & ACC   & NMI   & ACC   & NMI   & ACC   & NMI   & ACC   & NMI \\
    SSC   & 0.8631 & 0.8892 & 0.7191 & 0.8212 & 0.5510 & 0.5841 & 0.2441 & 0.2741 & 0.4530 & 0.4709 & 0.2189 & 0.1108 & 0.5741 & 0.6286 \\
    ENSC  & 0.8760 & 0.8952 & 0.7426 & 0.8380 & 0.5732 & 0.5924 & 0.2264 & 0.1689 & 0.4983 & 0.5495 & 0.1718 & 0.0503 & 0.6194 & \textbf{0.6578} \\
    SSC-OMP & 0.6410 & 0.7412 & 0.4431 & 0.6545 & 0.3271 & 0.6756 & 0.0764 & 0.0941 & 0.3400 & 0.3272 & 0.1115 & 0.0096 & 0.6148 & 0.6216 \\
    LRR   & 0.8118 & 0.8747 & 0.6493 & 0.7828 & 0.4682 & 0.4721 & 0.0676 & 0.0132 & 0.5386 & 0.5632 & 0.1188 & 0.0037 & 0.2463 & 0.1847 \\
    EDSC  & 0.8371 & 0.8828 & 0.6870 & 0.8139 & 0.6187 & 0.7369 & ——    & ——    & 0.5722 & 0.5823 & ——    & ——    & 0.4667 & 0.5794 \\
    DSC   & 0.9368 & 0.9408 & 0.8075 & 0.8941 & 0.6771 & 0.8908 & 0.1882 & 0.2496 & 0.7456 & 0.7420 & 0.1978 & 0.0718 & 0.5789 & 0.6215 \\
    DASC  & 0.9639 & 0.9686 & 0.8354 & 0.9196 & 0.7215 & 0.7286 & 0.1632 & 0.1475 & 0.8040 & 0.7800 & 0.2154 & 0.0768 & 0.5919 & 0.6147 \\
    S2ConvSCN & 0.9786 & 0.9633 & 0.6333 & 0.7516 & 0.7333 & 0.8522 & 0.0860 & 0.0700 & 0.8245 & 0.8134 & 0.1052 & 0.0080 & 0.4477 & 0.3865 \\
    Ours  & \textbf{0.9812} & \textbf{0.9826} & \textbf{0.9490} & \textbf{0.9894} & \textbf{0.7404} & \textbf{0.9440} & \textbf{0.5699} & \textbf{0.5391} & \textbf{0.9160} & \textbf{0.8981} & \textbf{0.4345} & \textbf{0.3341} & \textbf{0.6903} & 0.6337 \\
    \bottomrule
    \end{tabular} }%
\end{sidewaystable}%

In addition, to visually demonstrate the powerful performance of our models, we visualize the affinity matrix $W$ of S$^{3}$CE and DSC according to Eq. \ref{con:eq4} in Fig. \ref{fig:4}, where $W_{ij}$ represents the similarity between $Z_{i}$ and $Z_{j}$. Brighter pixel means higher similarity between two data points. As shown in the figure, most of the bright pixels are located at the block matrix on the diagonal. Compared with our method, DSC appears more bright pixels in non-diagonal regions, which means that the similarity between sample points calculated by the model is inaccurate. In addition, our method computes a clearer block-diagonal matrix, which means that the data are more evenly distributed. This partly explains the outstanding performance of our approach.

\begin{figure}[!h]
\centering
\includegraphics[scale=0.45]{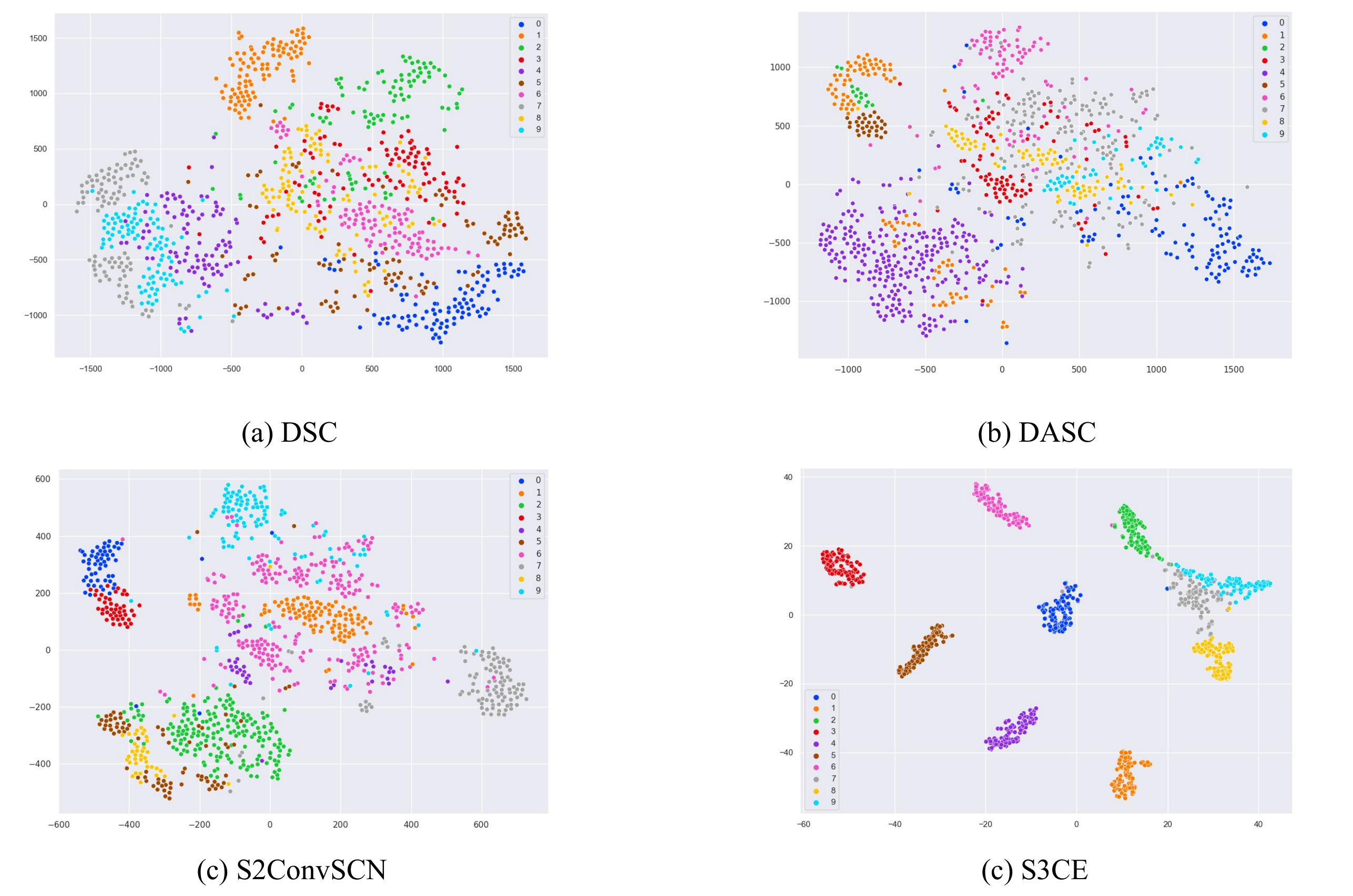}
\caption{2D visualization of the embedding spaces learned on the MNIST-1000 dataset}
\label{fig:5}
\end{figure}

Taking the MNIST-1000 data as an example, we visualize the output representations after a self-expression layer using the t-SNE \cite{bib40} method. As shown in Fig. \ref{fig:5}, the data point distribution has a serious overlap between clusters in DSC. This method can only separate the categories with more different features. For S2ConvSCN, this method introduces a classification module to supervise the training of feature learning, the inter-class distance of the method is enlarged. But the intra-class distance cannot be controlled to be a small value, so the effect is not too satisfactory. Because of the introduction of contrastive learning to extract features, it can be seen that the features extracted by the S$^{3}$CE method are not only discrete between classes but also compact within classes.

\begin{figure}[!h]
\centering
\includegraphics[scale=0.42]{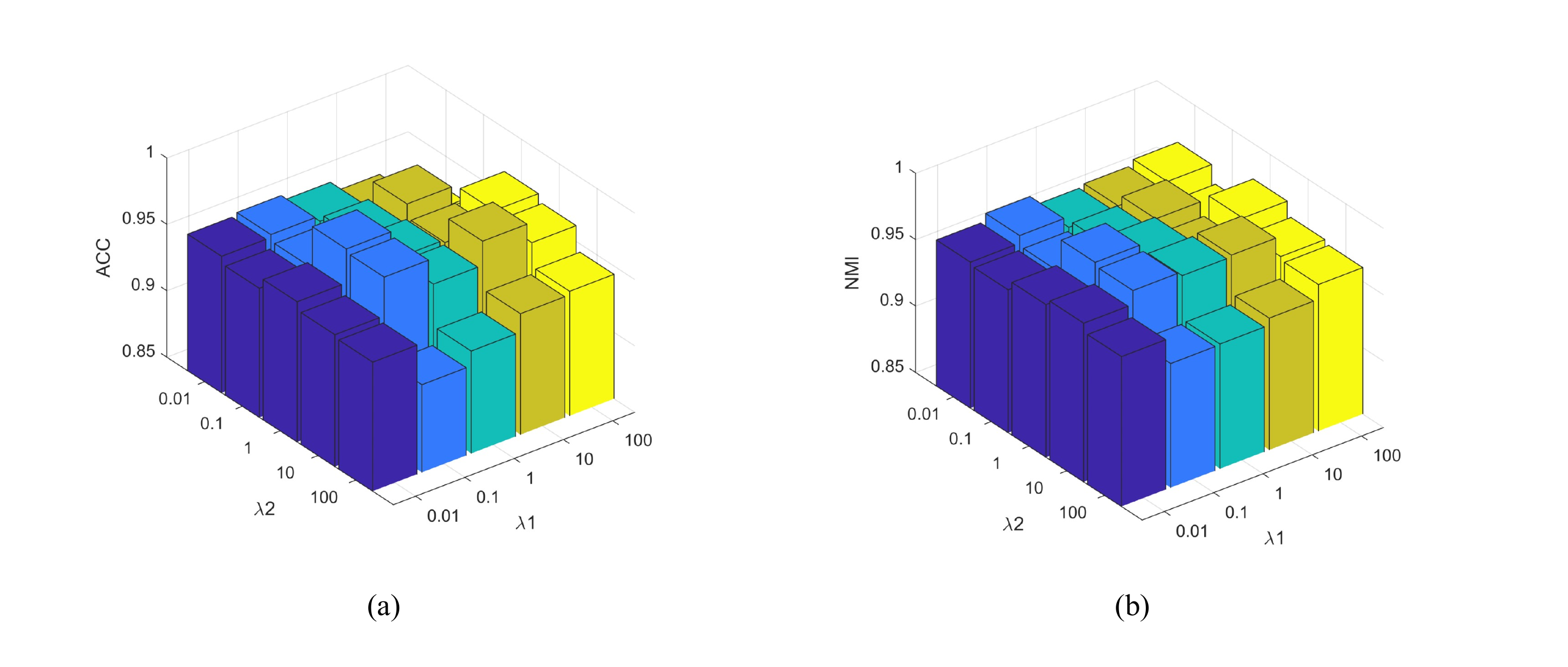}
\caption{The influence of parameters on ACC(a) and NMI(b) of COIL20 dataset}
\label{fig:6}
\end{figure}

\begin{figure}[!h]
\centering
\includegraphics[scale=0.42]{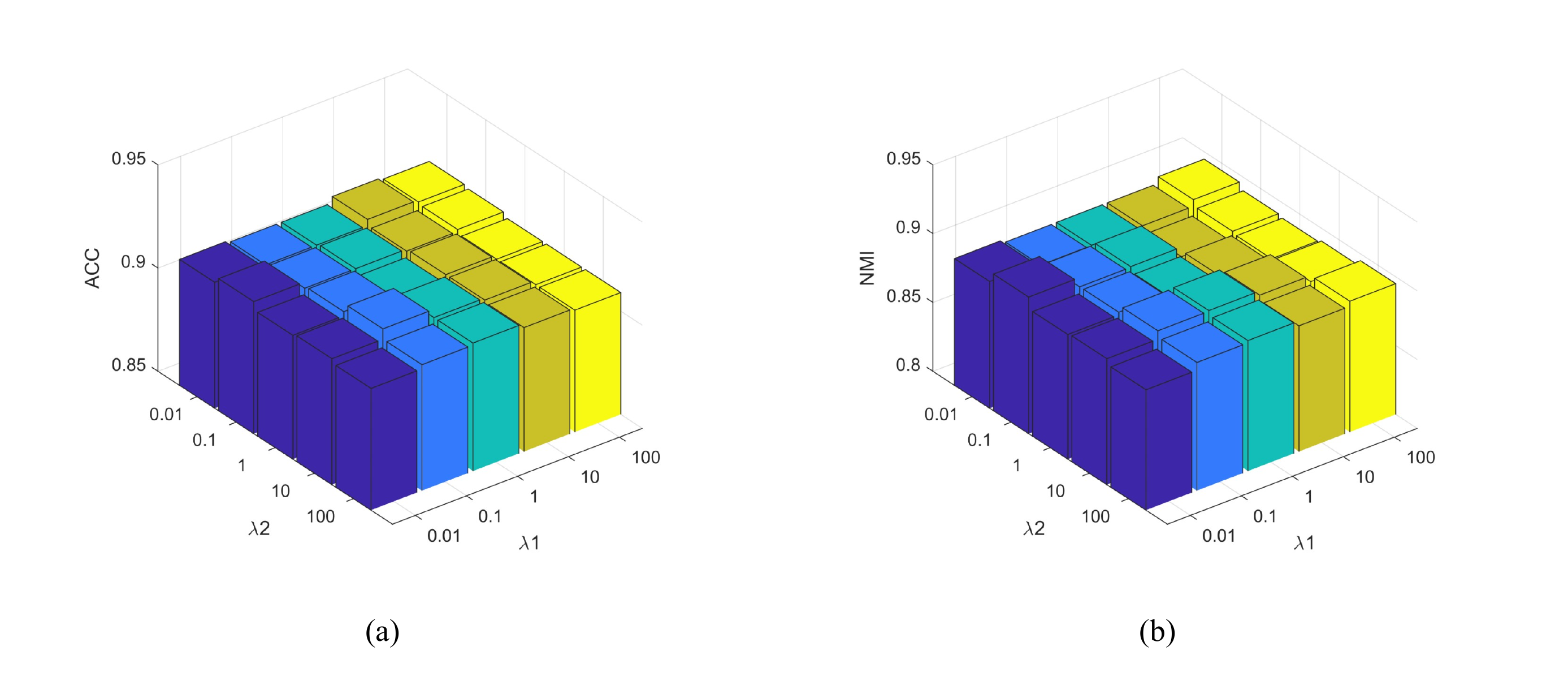}
\caption{The influence of parameters on ACC(a) and NMI(b) of MNIST-1000 dataset}
\label{fig:7}
\end{figure}

\subsection{Parameter Analysis}\label{subsec4}
There are two hyperparameters $\lambda_{1}$ and $\lambda_{2}$ in S$^{3}$CE. These two hyper-parameters affect the proportion of the self-expression loss and the entropy-norm in the total loss. In general, they are open and data-dependent, so we perform the sensitivity experiments on COIL20 and MNIST-1000. We can gain similar insight on the remaining datasets and thus leave out them. The ACC and NMI variation with the changes of $\lambda_{1}$ and $\lambda_{2}$ values are shown in Fig. \ref{fig:6}-\ref{fig:7} with three-dimensional histograms. Although the parameters affect the performance of our method, S$^{3}$CE outperforms within a wide range of values

\section{Conclusion}\label{sec5}
In this paper, we present a novel deep subspace clustering algorithm called S$^{3}$CE. S$^{3}$CE firstly combines the idea of contrastive leaning to obtain features with richer information while keeping the local structure information, and it introduces an extra regulation term entropy-norm that not only respects the intrinsic geometric but nonlocal structure of the latent representation. Experimental results on seven available datasets show that S3CE is superior to some start-of-the-art subspace clustering algorithms in clustering performance.

\bmhead{Acknowledgments}

This work was supported by Public-welfare Technology Application Research of Zhejiang Province in China under Grant LGG22F020032, and Key Research and Development Project of Zhejiang Province in China under Grant 2021C03137, Zhejiang Provincial Natural Science Foundation of China under Grant LY21F020001, Science and Technology Plan Project of Wenzhou in China under Grant ZG2020026.

\section*{Declarations}

The authors have no financial or proprietary interests in any material discussed in this article.


\bibliography{sn-reference}


\end{document}